\documentclass{article}
\usepackage[margin=0.75in]{geometry} %

\PassOptionsToPackage{numbers, compress}{natbib}

\usepackage[utf8]{inputenc} 
\usepackage[T1]{fontenc}    
\usepackage{hyperref}       
\usepackage{url}            
\usepackage{booktabs}       
\usepackage{amsfonts}       
\usepackage{nicefrac}       
\usepackage{microtype}      
\usepackage{xcolor}         
\usepackage{wrapfig}

\usepackage{bbm}
\usepackage[utf8]{inputenc} 
\usepackage[T1]{fontenc}    
\usepackage{hyperref}       
\usepackage{url}            
\usepackage{booktabs}       
\usepackage{amsfonts}       
\usepackage{nicefrac}       
\usepackage{microtype}      
\usepackage{algorithm}
\usepackage{algorithmic}
\usepackage{mathtools}
\usepackage{amsthm}
\usepackage{amssymb}
\usepackage{amsmath}
\usepackage{pifont}
\newcommand{\cmark}{\text{\ding{51}}}
\newcommand{\xmark}{\text{\ding{55}}}
\usepackage[capitalize,noabbrev]{cleveref}
\usepackage{thmtools, thm-restate}
\theoremstyle{plain}

\theoremstyle{definition}

\theoremstyle{remark}

\usepackage[textsize=tiny]{todonotes}
\usepackage{microtype}
\usepackage{graphicx}
\usepackage{subfigure}
\usepackage{natbib}
\usepackage{booktabs} 
\usepackage{enumitem}
\usepackage{titletoc}

\usepackage{wrapfig}
\usepackage[framemethod=TikZ]{mdframed}
\mdfsetup{%
    middlelinecolor =   none,
    middlelinewidth =   1pt,
    backgroundcolor =   blue!5,
    roundcorner     =   5pt,
}

\title{Supervised Fine-Tuning as Inverse Reinforcement Learning}

\author{%
  Hao Sun\thanks{hs789@cam.ac.uk. Preliminary thoughts and notes. Work in progress.} \\
  University of Cambridge\\
}

\begin{document}

\maketitle

\begin{abstract}
The prevailing approach to aligning Large Language Models (LLMs) typically relies on human or AI feedback and assumes access to specific types of preference datasets. In our work, we question the efficacy of such datasets and explore various scenarios where alignment with expert demonstrations proves more realistic. We build a sequential decision-making framework to formulate the problem of aligning LLMs using demonstration datasets. Drawing insights from inverse reinforcement learning and imitation learning, we introduce various approaches for divergence minimization in the LLM alignment tasks. Our analysis highlights the mass-covering and mode-seeking behaviors of these different approaches. Inclusively, we examine the pros and cons of the classical supervised fine-tuning method, elaborating on scenarios where different methods shine.
\end{abstract}

\section{Introduction}
While large language models (LLMs) alignment is a rapidly developing research area, the focus of existing research is mainly on reinforcement learning from human feedback (RLHF)~\citep{christiano2017deep,ouyang2022training} and their variants, e.g., deriving supervised learning objectives~\citep{rafailov2024direct}, applying contrastive learning~\citep{zhao2023slic}, introducing iterative supervised learning~\citep{yuan2023rrhf,dong2023raft}, regularizing the preference modeling~\citep{azar2023general}, or leveraging alternative approaches rooted in game theory~\citep{munos2023nash}.

Most of those approaches assume the existence of a preference dataset $\mathcal{D}_\mathrm{pref} = \{x_{i}, y_{i}^+, y_{i}^-\}_{i\in[N]}$ consisting $N$ pairwise preference information of language model responses $y^+$ (preferred) and $y^-$ (dispreferred) given query $x$. In general, such a dataset does not always exist, hence in most cases human annotators or advanced general-purpose LLMs are queried to provide annotation~\citep{bai2022constitutional,lee2023rlaif,guo2024direct}. Such a dataset can be extremely noisy sometimes~\citep{azar2023general} and \textbf{the underlying assumption of Bradley-Terry model~\citep{bradley1952rank} may rarely be satisfied} (We provide detailed analysis in Appendix~\ref{sec:bradley-terry}). Moreover, sharing data with annotators or commercial general-purpose LLMs may not be always possible, e.g., it can be restricted by privacy issues.
In comparison, human demonstrations or expert-crafted data under the format of $\mathcal{D}_\mathrm{exp} = \{x_i, y^*_i\}_{i\in[N]}$ is always of much higher quality. Yet in the literature, the usage of such a format of dataset is usually limited to supervised fine-tuning (SFT). In this work, we will \textbf{explore the potential of the SFT dataset from a formal reinforcement learning (RL) perspective}, providing rationales and empirical evidence of how to further exploit the SFT dataset in aligning LLMs.

\begin{mdframed}[innertopmargin=0pt,leftmargin=0pt, rightmargin=0pt, innerleftmargin=10pt, innerrightmargin=10pt, skipbelow=0pt]
\paragraph{Highlighted Take-Aways} 
\begin{enumerate}
    \item We argue that in LLM alignment, learning from demonstration can be more efficient than preference-based learning, especially when strong general-purpose LLM feedback is available.
    \item By formally defining the auto-regressive token generation as a sequential decision-making problem, we link the previous practice in RL with the context of LLM alignment. 
    \item With the formal definition, we study practical demonstration-based alignment algorithms from the perspective of RL. We show that the SFT objective is equivalent to trajectory-level distribution matching using the forward KL divergence, explaining their mass-covering behavior. 
    \item Furthermore, we discuss potential mode-seeking behaviors that other alignment approaches can provide using the reverse KL divergence or Jensen-Shannon divergence, and derive their practical objectives.
\end{enumerate}
\end{mdframed}


\section{Preliminaries}
\subsection{Markov Decision Processes}
RL can be formally represented using the Markov Decision Processes (MDPs), where decisions are made in discrete time steps, and each decision affects the state of the environment in the subsequent step.
Formally, an MDP can be denoted as $\mathcal{M} = \{\mathcal{S},\mathcal{A},\mathcal{T},\mathcal{R},\rho_0,\gamma\}$, where $\mathcal{S}\subset \mathbb{R}^{d}$ denotes the $d$-dim state space, $\mathcal{A}$ is the action space. Broadly, the environment includes $\mathcal{T}$ and $\mathcal{R}$, the former denotes the transition dynamics $\mathcal{T}: \mathcal{S}\times \mathcal{A} \mapsto \Delta(\mathcal{S})$ that controls transitions between states, and the reward function $\mathcal{R}:\mathcal{S}\times\mathcal{A}\mapsto \mathbb{R}$ provides feedback. $\rho_0 = p(s_0)\in\Delta(\mathcal{S})$ denotes the initial state distribution. $\gamma$ is the discount factor that trades off between short-term and long-term returns.

\subsection{Online and Offline RL}
\paragraph{Online RL} In the \textit{Online RL} setting, an agent with policy $\pi\in\Pi:\mathcal{S}\mapsto \Delta (\mathcal{A})$ learns through trial and error. It actively interacts with the environments --- including both transition dynamics $\mathcal{T}$ and the reward function $\mathcal{R}$. 

At each time step $t$, an agent observes a state $s_t$ from the environment and selects an action $a_t \sim \pi$. Upon taking the action, the agent receives a reward $r_t$ and transit to a new state $s_{t+1}$. The agent's objective is to maximize its expected return.
\begin{equation}
\label{eq:online-rl}
    \pi^* = \arg\max_{\pi\in\Pi}\mathbb{E}_{a_t\sim\pi, s_{t+1}\sim \mathcal{T},s_0\sim \rho_0}\sum_{t=0}^T \gamma^t \mathcal{R}(s_t, a_t),
\end{equation}

\paragraph{Offline RL}
In the \textit{Offline RL} setting, interactions with the environment are strictly forbidden. The learning problem is no longer online learning but learning from a static dataset of decision logs $\mathcal{D}_{\mathrm{Offline}} = \{(s^i_t,a^i_t,s^i_{t+1},r^i_t)\}$, that is generated by some unknown behavior policy $\pi_\beta$.

The most obvious difficulty in the offline RL setting is such a setting prohibits exploration --- hence it hinders the improvement of policy learning to be improved over the demonstration data.

\subsection{Behavior Clone and Imitation Learning}
\paragraph{Behavior Cloning (BC)}
Assuming the decision dataset is collected from an optimal behavior policy $\pi_\beta^*$, such that every decision $a^i_t$ is optimal. Denoting the state-action pairs in the dataset as $(s_t, a^*_t)$, the BC method learns a policy through a supervised learning objective that minimizes the difference between decision demonstration pairs. i.e.,
\begin{equation}
    \pi = \arg\min_\pi \mathbb{E}_{(s^i_t,a^i_t)\sim\mathcal{D}} ||a^i_t -\pi(s^i_t)||^2
\end{equation}
A fundamental challenge of BC is the \textit{distributional shift}: in evaluation, the state distribution is sampled from rolling out the learned policy $\pi$, rather than the behavior policy $\pi_\beta$ that generates the dataset.


\paragraph{Imitation Learning (IL)}

In order to alleviate the challenge of compounding error we discussed above, IL considers the setting where a dynamics model is available during learning. 
The objective of IL is to learn from a (decision) demonstration dataset, with access to a dynamics model --- such that the \textbf{current policy can be rolled out in the real environment}. Intuitively, with such a dynamics model, the optimization objective will no longer be $s_t\sim p_{\pi_\beta}(\tau)$ but could be $s_t\sim p_{\pi}(\tau)$ --- \textbf{the distributional shift problem can be alleviated.} It has been shown in the literature that having access to a \textit{dynamics model} is essential in controlling the error bound.~\cite{ross2011reduction}



\subsection{Reinforcement Learning from Human Feedback (RLHF)}
Introduced in the seminal paper of \cite{christiano2017deep}, RLHF provides an alternative to a scalar reward signal in reinforcing policy learning. In the LLM era, \cite{ouyang2022training} introduced the 3-step alignment framework for LLMs, namely the supervised fine-tuning (SFT), reward-modeling (RM), and policy learning with proximal policy optimization (PPO). Such a process assumes two different types of datasets: 1. the SFT dataset contains queries and expert-generated responses to those queries, under the form of $\mathcal{D}_\mathrm{exp} = \{x_i, y_i^*\}_{i\in[N_e]}$; and 2. the preference dataset $\mathcal{D}_\mathrm{pref} = \{x_i, y^+_i, y^-_i\}_{i\in[N_p]}$ that contains queries, multiple language model responses, and human preferences over those response labeled by human annotators.
The current practice of RLHF follows those two-stage and two-dataset frameworks, and several improvements were introduced in the literature: the DPO circumvents explicit reward modeling and stabilizes the learning process on preference dataset using supervised signals~\citep{rafailov2024direct}; SLiC-HF~\citep{zhao2023slic} gains insight from contrastive learning and learns from closed-form losses that maximize
the margin between the preferred and dispreferred generations; other alternatives include iterative supervised learning~\citep{yuan2023rrhf,dong2023raft}, regularizing the generation~\citep{azar2023general} or game-theory motivated methods~\cite{munos2023nash}.

\section{Rethinking LLM Alignment from an RL Perspective}
In this section, we will introduce our key insight that LLM alignment can be cast into the framework of forward and inverse RL. We first elaborate on the sequential decision-making nature of auto-regressive LLM generation in Section~\ref{sec:arllm-seqdecmak}; we then discuss the online nature and offline practice of LLM alignment in Section~\ref{sec:alignmentasonlinerl}; Finally, we highlight the perspective that LLM alignment can be formulated as an imitation learning problem, and introduce practical algorithms that circumvent the requirement of expensive preference data assumed in prevailing LLM alignment literature in Section~\ref{sec:alignmentwithimitation}.

\subsection{Auto-Regressive Language Generation as Sequential Decision Making}
\label{sec:arllm-seqdecmak}
In this section, we first cast the specific setting of auto-regressive language generation into the framework of MDP($\backslash$R).
In modern decoder-based architecture LLMs, we use $C$ to denote the context window size and use $\mathcal{V}$ to denote the vocabulary including the special tokens like \texttt{[EOS]} and \texttt{[MASK]}, the MDP is instantiated as follows: 
State space $\mathcal{S} = \mathcal{V}^C$; action space $\mathcal{A}=\mathcal{V}$; the transition dynamics is deterministic and known: $s' = \mathcal{T}(s,a) = \texttt{Concat}(s,a) = [s, a] $;  Specifically, we consider the states with an \texttt{[EOS]} token to be the absorbing states, i.e., $ \forall a: s' = \mathcal{T}(s,a|\texttt{[EOS]}\in s) = s$;
the LLM $\ell$ --- as a policy $\pi = \ell$ --- generates the next token $a\in\mathcal{A}$ based on the current context $s\in\mathcal{S}$; 

For instance, when the context window length is $C=6$, and an initial state $s_0$ is given as 
\begin{equation*}
    s_0 = \big[\texttt{ The | color | of | 
 the | sky |\hspace{1pt}[MASK]\hspace{1pt}|\hspace{1pt}[MASK]}\big].
\end{equation*}
when the language model policy $\pi$ selects a new token $\texttt{is}$ from the vocabulary $\mathcal{V}$, the next state deterministically becomes
\begin{equation*}
    s_1 = \texttt{Concate}(s_0, a_0=\texttt{is})= \big[\texttt{ The | color | of | 
 the | sky | is |\hspace{1pt}[MASK]}\big].
\end{equation*}
and the generation process continues with the language model policy, until when either an \texttt{[EOS]} token is selected or the maximal context window size is reached. The final generated context could be:
\begin{equation*}
    s_2 = \texttt{Concate}(s_1, a_1=\texttt{blue}) = \big[\texttt{ The | color | of | 
 the | sky | is | blue. }\big].
\end{equation*}

\subsection{Alignment as Online Reinforcement Learning with Human Feedback}
\label{sec:alignmentasonlinerl}
The research of LLM alignment studies how to align LLM with their users by generating responses that are more helpful, truthful, and harmless to the users~\cite{ouyang2022training}.

Under the framework of MDP, human users are the reward model $\mathcal{R}$ that provides feedback to the LLM generation process. 
In most cases, such an evaluation should be conducted on the whole response level, i.e., after the entire generation process is completed. 
\begin{equation}\label{eq:base_reward_function}
    \mathcal{R}(s_t,a_t) = \left\{
    \begin{array}{ll}
        r(s_t) & \text{if $s_t$ is a terminal state,}  \\
        0 & \text{otherwise}.
    \end{array}
\right.
\end{equation}
Ideally, human users could provide feedback to each response, and such signals can reinforce the language model generation through conventional RL algorithms. i.e., by solving the problem of Equation (\ref{eq:online-rl}). 

However, keep asking human users to provide feedback on language model-generated responses is unrealistic. Therefore, in practice, offline datasets are leveraged in the alignment process. In general, there are two types of data:

\paragraph{Offline Expert Demonstration (also known as the SFT Dataset)}
In real-world applications, the most general format of data that can be applied to align LLMs is the expert demonstration dataset $\mathcal{D}_\mathrm{exp} = \{x_i, y_i^*\}_{[i\in[N]}$. 
This data format is general, for instance, $x$ can be a general query for QA tasks, an incomplete sentence for completion tasks, or a general instruction of commands for instruction following tasks; and $y^*$ in those tasks are desired answers, completed sentence, or responses following the instruction, respectively.

In the literature, such a dataset is always used for supervised fine-tuning (SFT), hence such a format of dataset is also known as the SFT dataset. 
During SFT training, the learning objective is minimizing the token-wise difference given the existing context. For example, when 
\begin{equation*}
\begin{split}
    x_i &= \big[\texttt{ What | is | the | 
 color | of | the | sky? }\big], \\
    y^*_i &= \big[\texttt{ The | color | of | 
 the | sky | is | blue. }\big].
\end{split}
\end{equation*}
the SFT training first reorganizes the dataset as follows:
\begin{equation*}
\begin{split}
    s_0 &= \big[\texttt{ What | is | the | color | of | the | sky?~|\hspace{1pt}[MASK]\hspace{1pt}|\hspace{1pt}[MASK]\hspace{1pt}|\hspace{1pt}[MASK]\hspace{1pt}| ... }\big], \\
    a^*_0 &= \big[\texttt{ The }\big], \\
    s_1 & = \big[\texttt{ What | is | the | 
 color | of | the | sky?~| The |\hspace{1pt}[MASK]\hspace{1pt} |\hspace{1pt}[MASK]\hspace{1pt}| ... }\big], \\
    a^*_1 &= \big[\texttt{ color }\big], \\
    s_2 & = \big[\texttt{ What | is | the | 
 color | of | the | sky?~| The | color |\hspace{1pt}[MASK]\hspace{1pt}| ... }\big], \\
    a^*_2 &= \big[\texttt{ of }\big], \\
    &...
\end{split}
\end{equation*}
with such a dataset, the learning objective is to reproduce demonstration token $a^*_j $ when feeding $s_j$ to the LLM (policy). The training of the SFT is conducted through supervised classification.

\paragraph{Offline Preference Data} Another type of data, which is widely studied in the literature, is the preference dataset labeled by human annotators $\mathcal{D}_\mathrm{pref} = \{x_i, y_i^+, y_i^-\}_{i\in[N]}$. In such a dataset, multiple responses are generated by the LLM policy --- rather than generated by human experts --- and then ranked by human annotators. The general assumption made behind using such a dataset is that preference data is much easier and cheaper to collect. Moreover, ranking different generated contexts can be much easier than directly providing a score for each of those responses.

\subsection{Imitation Learning for Alignment with Offline Feedback}
\label{sec:alignmentwithimitation}
In this section, we argue that LLM alignment with the \textbf{offline feedback dataset} can be formulated as an \textbf{online-IL} problem. 

At the first glance, the LLM alignment with an offline dataset might seem to be an offline RL problem, in the sense that no more interactions with the human annotators are available during training. However, in RL literature, the accessibility of online interactions with the \textbf{dynamics model}, rather than the \textbf{reward model}, determines the online or offline nature of the tasks. In LLM alignment, while it is impossible to access the reward models (human annotators) during training,  \textbf{\textit{the dynamics model in response generation is known and accessible.}} Specifically, the actions are tokens generated by LLMs, and the responses (trajectories) are concatenations of those generated tokens.

Practically, RLHF chooses to use the Inverse RL approach for the IL problem --- with the first step explicitly learning a reward model, and the second step conducting RL using the known dynamics model and the learned reward model. However, converting a preference-based dataset into a reward model requires non-trivial effort~\cite{rafailov2023direct}, and the assumptions under the prevailing Bradley-Terry model can hardly be satisfied in practice~\cite{azar2023general,munos2023nash}.

On the other hand, in conventional research of RL, learning from human feedback through preference learning is not the only choice. Learning from expert demonstration has been widely applied to robotics control~\cite{schaal1996learning,nair2018overcoming,hester2018deep}, autonomous driving vehicles~\cite{kuderer2015learning,scheel2022urban}, playing video game~\cite{vinyals2019grandmaster}, and AlphaGo~\cite{silver2016mastering}. We contrast the differences between RL, Offline-RL, IL, Offline-IRL, and Learning from Demonstration (LfD) problem settings in Table~\ref{tab:alihan-dan}.

\begin{table}[h]
\fontsize{8}{10}\selectfont
\centering
\caption{\small Summarizing difference in problem settings of RL, Offline-RL, IL, IRL, Offline-IRL, and LfD.}
\begin{tabular}{l|c|c|c|c|c}
\toprule
\textbf{Problem} & \textbf{External} & \textbf{External} & \textbf{Learned} & \textbf{Demonstration} & \textbf{Examples} \\
\textbf{Settings} & \textbf{Dynamics} & \textbf{Reward} & \textbf{ Reward} &  & \textbf{} \\
\textbf{} & \textbf{Model} & \textbf{Model} & \textbf{Model} & \textbf{} & \\
\midrule
RL & $\cmark$ & $\cmark$ & $\xmark$ & $\xmark$ & PPO~\cite{schulman2017proximal}, TD3~\cite{fujimoto2018addressing},SAC~\cite{haarnoja2018soft}\\
Offline-RL & $\xmark$ & $\xmark$ & $\cmark$ or $\xmark$ & $\cmark$ & BC~\cite{pomerleau1991efficient}, AOC~\cite{sun2023accountable}, CQL~\cite{kumar2020conservative}, WGCSL~\cite{yang2022rethinking} \\
\textbf{IL} & $\cmark$ & $\xmark$ & $\cmark$ or $\xmark$ & $\cmark$ & BC~\cite{pomerleau1991efficient}, AOC~\cite{sun2023accountable}, GAIL~\cite{ho2016generative} \\
\textbf{IRL} & $\cmark$ & $\xmark$ & $\cmark$ & $\cmark$ & BC~\cite{pomerleau1991efficient}, AOC~\cite{sun2023accountable}, T-REX~\cite{brown2019extrapolating}, AIRL~\cite{fu2017learning} \\
Offline-IRL & $\xmark$ & $\xmark$ & $\cmark$ & $\cmark$ & BC~\cite{pomerleau1991efficient}, AOC~\cite{sun2023accountable}, SBIL~\cite{jarrett2020strictly} \\
LfD & $\cmark$ & $\cmark$ & $\xmark$ & $\cmark$ & DQNfD~\cite{hester2018deep}, DDPGfD~\cite{nair2018overcoming}, AlphaStar~\cite{vinyals2019grandmaster} \\
\bottomrule
\end{tabular}
\label{tab:alihan-dan}
\end{table}

\subsection{Alignment as Inverse RL: from Behavior Cloning to Adversarial Imitation}
\label{sec:imitation}
Instead of following the prevailing approaches in the LLM alignment research where a preference dataset is utilized, in this work, we focus on the offline expert demonstration dataset which is more accessible in real-world applications. And aim at developing algorithms for LLM alignment based on such a dataset that can surpass the performance of SFT --- the common practice on such a dataset.

The usage of the demonstration dataset, together with the accessibility of the dynamics model, posit the problem naturally as an IL task. In literature, the simplest approach to IL is Behavior Cloning~\cite{pomerleau1991efficient}, which leverages supervised learning to predict the actions in the demonstration dataset given the states. It was shown in the literature such an action-space similarity matching is unreliable due to the compounding errors~\cite{ross2011reduction}.
Adversarial Imitation Learning (AIL) algorithms~\cite{ho2016generative,fu2017learning,ghasemipour2020divergence,kostrikov2018discriminator,orsini2021matters} solve the problem by gaining inspiration from both Generative Adversarial Networks (GANs)~\cite{goodfellow2014generative} and Inverse RL~\cite{ng2000algorithms,ziebart2008maximum}: starting from the objective of distributional matching, GAIL aims at learning a policy whose \textbf{state-action space occupancy measure} is indistinguishable from the occupancy measure of the expert demonstrations.

We denote the state-action occupancy measure of the behavior policy as $\rho^\mathrm{exp}(s,a) = \pi_\mathrm{exp}(a|s)\sum_{t=0}\gamma^t P(s_t = s|\pi_\mathrm{exp})$, and denote the state-action occupancy measure of the current policy as $\rho^\pi(s,a)$. Intuitively, the occupancy measure describes the distribution of state-action pairs that an agent visits in the space with policy.
For auto-regressive LLMs which take context $x$ as input and output response $y = (y^{(0)},y^{(1)},...,y^{(K)}=\texttt{EOS} )$, we have 
\begin{equation}
    \rho^\pi(s_k,a_k) = \rho^\pi(s_k = (x,y^{(0:k-1)}),a_k = y^{(k)}) = p(x)\Pi^{t=k}_{t=0} \pi(a_t = y^{(t)}| s_t = (x,y^{(0:t-1)}))
\end{equation}
In addition, we denote the trajectory distribution as the occupancy measure of completed generations 
\begin{equation}
    d^\pi(y|x)=\Pi^{t=K}_{t=0} \pi(a_t = y^{(t)}| s_t = (x,y^{(0:t-1)})) = \rho^\pi(s_{\textcolor{red}{K}},a_{\textcolor{red}{K}})/p(x)
\end{equation}
We can use the demonstration dataset to approximate the trajectory distribution of the expert policy:
\begin{equation}
    d^\mathrm{exp}(y|x)\approx P( (x,y)\in\mathcal{D}_\mathrm{exp}|x\in\mathcal{D}_\mathrm{exp})
\end{equation}

In the following, we link different practical learning objectives in the Inverse RL literature and derive task-specific objectives in the context of LLM alignment.

\textcolor{brown}{\subsubsection{Alignment with SFT --- Behavior Cloning}}
The learning objective of SFT is to minimize the negative log-likelihood of generating expert-generated tokens given the existing context 
\begin{equation}
\label{eqn:7}
    \min_\pi \mathbb{E}_{(s,a)\sim\rho^\mathrm{exp}} \left[\mathrm{KL}(\pi^\mathrm{exp}(a|s)||\pi(a|s)) \right] = - \max_\pi \mathbb{E}_{(s,a)\sim\rho^\mathrm{exp}} \left[\log(\pi(a|s)) \right]
\end{equation}
Therefore, the conventional SFT training objective minimizes the KL divergence of \textbf{action marginal distribution} between the behavior policy $\pi^\mathrm{exp}$ and the current policy $\pi$.

\textcolor{brown}{\subsubsection{Alignment with the Forward KL-Divergence}}
When minimizing the forward KL divergence between \textbf{state-action occupancy measures}
\begin{align}
    \min_\pi \left[\mathrm{KL}(\rho^\mathrm{exp}(s,a)||\rho^\pi(s,a)) \right] &= - \max_\pi \mathbb{E}_{(s,a)\sim\rho^\mathrm{exp}} \left[\log \rho^\pi(s,a) \right] \\
    &= - \max_\pi \mathbb{E}_{(s_k,a_k)\sim\rho^\mathrm{exp}} \left[\log \Pi^k_{t=0} \pi(a_t|s_t) \right] \\
    &= - \max_\pi \mathbb{E}_{(s_k,a_k)\sim\rho^\mathrm{exp}} \left[\sum^k_{t=0}\log \pi(a_t|s_t) \right] \\
    &= - \max_\pi \mathbb{E}_{(s_k,a_k)\sim \textcolor{red}{\widetilde{\rho^\mathrm{exp}}} } \left[\log \pi(a_k|s_k) \right] \\
    &= - \max_\pi \mathbb{E}_{(s_k,a_k)\sim \rho^\mathrm{exp}} 
 \left[ \textcolor{red}{\frac{K-k}{K}}\log \pi(a_k|s_k) \right]
    \label{eqn:12}
\end{align}
When minimizing the forward KL divergence between \textbf{trajectory distributions},
\begin{align}
    \min_\pi \left[\mathrm{KL}(d^\mathrm{exp}(y|x)||d^\pi(y|x)) \right] &= - \max_\pi \mathbb{E}_{(x,y)\sim \mathcal{D}^\mathrm{exp}} \left[\log d^\pi(y|x) \right] \\
    &= - \max_\pi \mathbb{E}_{(x,y^{(0:K)})\sim\mathcal{D}^\mathrm{exp}} \left[\sum^{K}_{t=0}\log \pi(a_t|s_t) \right]
    \label{eqn:16}
\end{align}
\begin{mdframed}[innertopmargin=0pt,leftmargin=0pt, rightmargin=0pt, innerleftmargin=10pt, innerrightmargin=10pt, skipbelow=0pt]
\paragraph{Take-Aways} 
    Comparing Equation (\ref{eqn:16}), Equation (\ref{eqn:12}), and Equation (\ref{eqn:7}), we can conclude the following:\\ 1. Minimizing action marginal distribution between the demonstration dataset and the current policy leads to the SFT learning objective. \\
    2. Minimizing the forward KL divergence of \textbf{trajectories} between demonstration and current policy leads to the same learning objective as SFT. Yet it hints at seeking a sequential rather than random sampling method during training. \\
    3. Minimizing the forward KL divergence of \textbf{state-action occupancy measure} is different from the SFT objective by a re-weighting factor, depending on the \textbf{position of the token in the demonstration sequence}. Intuitively, it can be understood as a re-weighting approach to avoid compounding errors. \\
    4. As it is known that using the forward KL-Divergence will lead to mass-covering and using reverse KL-Divergence leads to mode-seeking behaviors~\cite{ghasemipour2020divergence,khalifa2020distributional,wiher2022decoding,wang2023beyond}, the approaches above are all mass-covering given their equivalences. \textbf{As a consequence, those SFT-type objectives are more suitable for close-ended tasks.}
    
\end{mdframed}

\textcolor{brown}{\subsubsection{Alignment with the Reverse KL-Divergence}}
 In the pursuance of mode-seeking behavior, we can minimize the Reverse KL divergence. When considering the reverse KL divergence on the \textbf{state-action occupancy measure}, the learning objective is
\begin{equation}
\begin{split}
\label{eqn:reverse-KL}
    \min_\pi [\mathrm{KL}(\rho^\pi(s,a)||\rho^\mathrm{exp}(s,a))] = -\max_\pi \mathbb{E}_{(s,a)\sim\rho^\pi}\left[ \log \rho^\pi(s,a) - \log \rho^\mathrm{exp}(s,a) \right].
\end{split}
\end{equation}
The difficulty in the above learning objective is that the second term is always unknown. In the literature, such a difficulty has been solved through adversarial training~\cite{fu2017learning}. By training a discriminative model $D_\phi$ parameterized by $\phi$, that learns to classify state-actions sampled from the demonstration dataset or from the behavior policy $\pi$, we get 
\begin{equation}
\label{eqn:optimal_d}
    D^*_\phi(s,a) = \frac{\rho^\mathrm{exp}(s,a)}{\rho^\mathrm{exp}(s,a)+\rho^\pi(s,a)}
\end{equation} 
at its optimal convergence~\cite{goodfellow2014generative}. 
Plugging Equation~(\ref{eqn:optimal_d}) into Equation~(\ref{eqn:reverse-KL}), an practical policy learning objective can be given by
\begin{equation}
    \min_\pi \mathbb{E}_{(s,a)\sim\rho^\pi}\left[ \log D_\phi(s,a) - \log (1-D_\phi(s,a)) \right]
\end{equation}
and $D_\phi$ is optimized iteratively through:
\begin{equation}
    \max_\phi \mathbb{E}_{(s,a)\sim\rho^\mathrm{exp}}[\log D_\phi(s,a)] +  \mathbb{E}_{(s,a)\sim\rho^\pi}[\log (1-D_\phi(s,a))]
\end{equation}
If we instead minimize the reverse KL divergence between the \textbf{trajectory distributions}, the policy learning objective and discriminator learning objective will become
\begin{equation}
    \min_\pi \mathbb{E}_{(y|x)\sim d^\pi}\left[ \log D_\phi(y|x) - \log (1-D_\phi(y|x)) \right]
\end{equation}
and 
\begin{equation}
    \max_\phi \mathbb{E}_{(y|x)\sim d^\mathrm{exp}}[\log D_\phi(y|x)] +  \mathbb{E}_{(y|x)\sim d^\pi}[\log (1-D_\phi(y|x))]
\end{equation}
respectively.

\textcolor{brown}{\subsubsection{Alignment with the Jensen–Shannon Divergence}}
Similarly, if we choose $f$ to be the Jensen-Shannon divergence, and minimize the divergence between \textbf{state-action occupancy measure},
\begin{equation}
\begin{split}
    & \min_\pi D_{JS}(\rho^\pi(s,a) || \rho^\mathrm{exp}(s,a)) \\
    & = \min_\pi \frac{1}{2}\mathrm{KL}\left(\rho^\pi(s,a)  \Bigg|\Bigg| \frac{\rho^\mathrm{exp}(s,a) + \rho^\pi(s,a)}{2}\right) + \frac{1}{2}\mathrm{KL}\left(\rho^\mathrm{exp}(s,a) \Bigg|\Bigg| \frac{\rho^\mathrm{exp}(s,a) + \rho^\pi(s,a)}{2}\right) \\
    & = \min_\pi \mathbb{E}_{(s,a)\sim\rho^\mathrm{exp}(s,a)}\left[\log D^*_\phi(s,a) \right] + \mathbb{E}_{(s,a)\sim\rho^\pi}\left[\log(1-D^*_\phi(s,a))\right],
\end{split}
\end{equation}
where $D^*_\phi(s,a) = \frac{\rho^\mathrm{exp}(s,a)}{\rho^\mathrm{exp}(s,a)+\rho^\pi(s,a)}$ is the optimal discriminator~\cite{goodfellow2014generative}. Practically, such an objective can be optimized by solving the following minimax game~\cite{ho2016generative,fu2017learning}:
\begin{equation}
    \min_\pi \max_{\phi} \mathbb{E}_{(s,a)\sim\rho^\mathrm{exp}}\left[\log D_\phi(s,a) \right] + \mathbb{E}_{(s,a)\sim\rho^\pi}\left[\log(1-D_\phi(s,a))\right],
\end{equation}

On the other hand, if we minimize the Jensen-Shannon divergence between the \textbf{trajectory distribution} $D_\mathrm{JS}(d^\pi(y|x)||d^\mathrm{exp}(y|x))$, the practical learning objective is 
\begin{equation}
    \min_\pi \max_{\psi} \mathbb{E}_{(y|x)\sim d^\mathrm{exp}}\left[\log D_\psi(y|x) \right] + \mathbb{E}_{(y|x)\sim\rho^\pi}\left[\log(1-D_\psi(y|x))\right],
\end{equation}
\begin{mdframed}[innertopmargin=0pt,leftmargin=0pt, rightmargin=0pt, innerleftmargin=10pt, innerrightmargin=10pt, skipbelow=0pt]
\paragraph{Take-Aways} 
    Comparing the learning objectives we derived when using the reverse KL divergence or the Jensen-Shannon divergence to the SFT-type of objectives above,  \\
    1. Performing mode-seeking is generally harder than mass-covering, which is caused by the \textbf{difficulty of estimating the probability of getting on-(current)-policy actions with the expert policy}. \\
    2. Such a difficulty can be circumvented through adversarial training. In general, there are two choices for learning the discriminative model corresponding to identifying the state-action occupancy measure and the trajectory distribution, respectively. \\
    3. Different from the SFT-type of learning objectives, the adversarial learning approaches do not only seek mass-covering. The superiority of such a class of approaches has been demonstrated in the low demonstration data regime~\cite{ghasemipour2020divergence}. \textbf{Consequently, the adversarial learning approaches are more suitable for open-ended tasks, especially under the low-demonstration regime.}
    
\end{mdframed}

\textcolor{brown}{\subsubsection{Discussion on DPO: the Reward Ambiguity and the Bradley-Terry Assumption}}
It is worth noting the link and differences with Direct Preference Optimization (DPO)~\cite{rafailov2023direct} and their self-play counterparts designed for alignment with the demonstration dataset~\cite{chen2024self}. Regardless of the data format DPO-type algorithms work on, the most important difference is that DPO explicitly \textbf{assumes the existence of a score-based scalar reward based on the Bradley-Terry model.}

Conversely, adversarial learning approaches utilizing discriminative models do not hinge on such explicit assumptions. While limiting the reward function to a specific function class may alleviate the reward ambiguity issue in inverse RL~\cite{fu2017learning,ng2000algorithms,ng1999policy,chan2024dense}, it also reduces the expressivity of the reward space. 

The primary objective in aligning Large Language Models is to generate high-quality responses, focusing less on recovering the exact reward function that demonstrators optimize for. Ideally, one should aim to directly learn the optimal policy for response generation without explicitly modeling the reward function. While DPO \textbf{sidesteps the need to explicitly parameterize such a reward model, it still relies on the Bradley-Terry model's assumption and the existence of a reward model}. In contrast, the adversarial imitation approaches introduced in our work do not presuppose any specific reward model form. They conceptually allow for a wider range of alternatives to the Bradley-Terry model, including direct preference objectives~\cite{azar2023general,munos2023nash} and prospect theory objective~\cite{ethayarajh2024kto}.



\section{Conclusive Remark}

This paper presents a novel approach to Large Language Model (LLM) alignment, utilizing insights from adversarial imitation learning. We conceptualize auto-regressive LLM generation as a sequential decision-making process within a Markov Decision Process framework.

Our investigation reveals that Supervised Fine-Tuning (SFT) objectives in LLMs align with trajectory-level distribution matching characterized by the forward KL divergence. This theoretical underpinning clarifies the mass-covering behavior inherent in these models. Additionally, we explore alternative alignment strategies employing reverse KL divergence or Jensen-Shannon divergence. These methods hint at potential mode-seeking behaviors and offer practical objectives for implementing these approaches.

\newpage
\bibliography{reference}
\bibliographystyle{unsrt}
\newpage
\appendix

\section{Assumptions behind Explicit Reward Modeling: the Bradley-Terry Model and Its Alternatives}
\label{sec:bradley-terry}



The Bradley-Terry model~\citep{bradley1952rank} and Elo score~\citep{elo1978rating} were originally developed for rating chess players, where the pairwise competition logs are switched to absolute scores. 

\textbf{The Gaussian Assumption on Performance}
To be specific, the Bradley-Terry model assumes the ability of \textbf{players} can be expressed as a score. In each two-player game, each player's performance will be a Gaussian distribution centered at this score. The variances of those Gaussian distributions are induced by the stochastic nature of the game, and variability of the players' performance. 

For instance, when player $A$ having score $S_A$ variance $\sigma_A$ and player $B$ having score $S_B$ variance $\sigma_B$ are playing against each other in a game, the probability that $A$ wins $B$ ($A \succ B$) in a game given the above Gaussian assumption on performance gives the following result:
\begin{equation}
    P(A \succ B) =P\left(x_a \ge x_b | x_a\sim N(S_A, \sigma_A^2), x_b\sim N(S_B, \sigma_B^2)\right) = \frac{1}{2} + \frac{1}{2}\mathrm{erf}\left(\frac{S_A-S_B}{\sqrt{2(\sigma_A^2+\sigma_B^2)}}\right)
\end{equation}
In practice, other sigmoid-type functions besides the error function $\mathrm{erf}(\cdot)$ can be used, e.g., using $\mathrm{tanh}(\cdot)$ when assuming the distribution is logistic.

\textbf{Bradley-Terry Model in LLM Alignment}
When it comes to RLHF, the Bradley-Terry model is applied to transfer \textbf{preference-based data} into scores. In such a process, the human evaluation is noisy and the probability of observing response $y_A$ to be preferred over response $y_B$ is expressed as 
\begin{equation}
    P(y_A \succ y_B|x) = \frac{1}{2} + \frac{1}{2}\mathrm{tanh}\left(\frac{r_A-r_B}{\sqrt{2(v_A^2 + v_B^2)}}\right)
\end{equation}
where $v_A, v_B$ models the variation in evaluating the value of different responses, and $r_A, r_B$ are the corresponding standardized scores of response $y_A, y_B$ given query $x$, respectively. 

In principle, there are two functions to be estimated given a preference dataset $\mathcal{D}_\mathrm{pref} = \{x_{i}, y_{i}^+, y_{i}^-\}_{i\in[N]}$. 
\begin{enumerate}
    \item First, the reward function $R_\theta:\mathcal{X}\times \mathcal{Y}\mapsto \mathbb{R}$ evaluates how good an answer $y\in\mathcal{Y}$ is for a query $x\in\mathcal{X}$. e.g., $r_A = R_\theta(x,y_A)$, $r_B = R_\theta(x,y_B)$. 
    \item Second, the variation function $V_\phi:\mathcal{X}\times \mathcal{Y}\mapsto \mathbb{R}$ evaluates how hard it is to evaluate whether an answer $y\in\mathcal{Y}$ is for a query $x\in\mathcal{X}$ is better than the other. e.g., $v_A = V_\phi(x,y_A)$, $v_B = V_\phi(x,y_B)$. 
\end{enumerate}
Using the Cross-Entropy Loss to fit $\mathcal{D}_\mathrm{pref}$, we have
\begin{equation}
\mathcal{L}_\mathrm{CE} = -\mathbb{E}_{(x,y^+,y^-) \sim \mathcal{D}_\mathrm{pref}}\left[\log \sigma 
\left(\frac{R_\theta(x, y^+) - R_\theta(x, y^-)}{\sqrt{(V^2_\phi(x, y^+) + V^2_\phi(x, y^-))/2}} \right)\right]
\end{equation}

In the common practice of RLHF based on the Bradley-Terry model~\citep{christiano2017deep,ouyang2022training,rafailov2024direct}, the learning of reward model only focuses on the score and eliminates the variation in evaluation. Therefore, the denominator is \textbf{simplified by setting} $V^2_\phi(x, y^+) = V^2_\phi(x, y^-) = 1$, i.e., the score is normalized by the variation of the problem

\begin{equation}
\widetilde{\mathcal{L}}_\mathrm{CE} = -\mathbb{E}_{(x,y^+,y^-) \sim \mathcal{D}_\mathrm{pref}}\left[\log \sigma 
\left(R_\theta(x, y^+) - R_\theta(x, y^-)\right)\right]
\end{equation}

The Bradley-Terry model in RLHF assumes human annotators’ preference can be expressed as scores centered at the real scores of different responses, yet it differs from the Bradley-Terry model used in chess rating or games in the sense that
\begin{enumerate}
    \item The RLHF dataset contains queries from different domains, some of which are intrinsically harder to evaluate, hence directly using the B-T model is to some extent like using a unified rating system of chess, Go, and poker --- the scores are not well calibrated.
    \item Different from chess, where the \texttt{number of players} $\ll$ \texttt{number of games}, in RLHF, the number of players (query-response pairs) is comparable to the number of games (annotator comparison).
    \item The Elo scores are executed and updated in an online manner, and offline learning with preference-based data may lose the ability to error correction.
Among those challenges, (1) and (2) can potentially be addressed with a learned variance term in the B-T model.
\end{enumerate}




\section{The General Framework $f$-Divergence}
Formally, according to the $f$-divergence framework of GANs~\cite{nowozin2016f} and Inverse RL~\cite{ghasemipour2020divergence}, the alignment problem can be written as training an LLM model $\pi$, such that
\begin{equation}
\label{eqn:fgan}
    \min_{\pi} \max_{T_\omega} \mathbb{E}_{(s,a)\sim\mathcal{D}_\mathrm{exp}}[T_\omega(s,a)] - \mathbb{E}_{(s,a)\sim\pi}[f^*(T_\omega(s,a))]
\end{equation}
 where $f:\mathbb{R}^+\mapsto\mathbb{R}$ is a convex, lower-semicontinuous
function, and it defines a statistical divergence between distribution $P,Q$ with density function $p,q$ as: $D_f(P||Q) = \large\int_x q(x) f\left(\frac{p(x)}{q(x)}\right)dx$, and $f^*$ is the conjugate of $f$, defined as $f^* = \sup_{u\in \mathrm{dom}_f}\{ut - f(u)\}$. Practically, it was shown in \cite{ghasemipour2020divergence} that Equation (\ref{eqn:fgan}) can be solved through iterative optimizing
\begin{equation}
\label{eqn:fgan-inner}
    \max_{T_\omega} \mathbb{E}_{(s,a)\sim\mathcal{D}_\mathrm{exp}}[T_\omega(s,a)] - \mathbb{E}_{(s,a)\sim\pi}[f^*(T_\omega(s,a))]
\end{equation}
and
\begin{equation}
\label{eqn:fgan-inner}
    \max_{\pi} \mathbb{E}_{\tau\sim\pi}[\sum_{t}f^*(T_\omega(s_t,a_t))]
\end{equation}
To elaborate on how different choices of $f$ lead to different practical implementations of the AIL approach of alignment, we take the state-action occupancy measure here for example:
\begin{itemize}
    \item AIRL: $f(u) = - \log (u)$ ; ~~\qquad\qquad \qquad\qquad$D_f(\rho^\mathrm{exp} || \rho^\pi) = \mathrm{KL}(\rho^\pi || \rho^\mathrm{exp}) $
    \item GAIL: $f(u) = -(u + 1) \log \frac{1+u}{2} +  u \log u$; \quad$D_f(\rho^\mathrm{exp} || \rho^\pi) = \mathrm{JS}(\rho^\pi|| \rho^\mathrm{exp} ) $
    \item FAIRL: $f(u) = u\log (u)$; ~~\qquad\qquad\qquad\qquad$D_f(\rho^\mathrm{exp} || \rho^\pi) = \mathrm{KL}(\rho^\mathrm{exp} || \rho^\pi)$
    \item $\alpha$-IRL: $f(u) = \frac{u^{1-\alpha} - (1-\alpha)u -a}{\alpha(\alpha-1)}$; ~~\quad\qquad\qquad $D_f(\rho^\mathrm{exp} || \rho^\pi) = D_\alpha(\rho^\mathrm{exp} || \rho^\pi)$
\end{itemize}
Therefore, the methods discussed in the main context could be extended to other divergences in the $f$-Divergence framework.
\end{document}